\documentclass[10pt,twocolumn,letterpaper]{article}

\usepackage{wacv}
\usepackage[pagebackref=true,breaklinks=true,letterpaper=true,colorlinks,bookmarks=false]{hyperref}
\usepackage{times}
\usepackage{epsfig}
\usepackage{graphicx}
\usepackage{subfig}
\usepackage{amsmath}
\usepackage{amssymb}
\usepackage[capitalize,noabbrev]{cleveref}
\usepackage{color}
\usepackage{verbatim}

\wacvfinalcopy % *** Uncomment this line for the final submission
\ifwacvfinal\fi
\begin{document}

%%%%%%%%% TITLE
\title{Saliency Prediction for Mobile User Interfaces}

% Authors at the same institution
%\author{First \hspace{2cm} Author \\
%Instituti\\
%{\tt\small fedauthor@adobe.com}
%}
% Authors at different institutions
\author{Prakhar Gupta \\
Adobe Research\\
{\tt\small prakhgup@adobe.com}
\and
Shubh Gupta\thanks{These three authors contributed equally} \\
IIT Kanpur\\
{\tt\small shubhg1996@gmail.com}
\and
Ajaykrishnan Jayagopal\footnotemark[1] \\
IIT Madras\\
{\tt\small akj1996@gmail.com}
\and
Sourav Pal\footnotemark[1] \\
IIT Kharagpur\\
{\tt\small souraviitcse.2015@gmail.com}
\and
Ritwik Sinha \\
Adobe Research\\
{\tt\small risinha@adobe.com}
}

\maketitle
\ifwacvfinal\thispagestyle{empty}\fi

%%%%%%%%% ABSTRACT
\begin{abstract}

We introduce models for saliency prediction for mobile user interfaces. A mobile interface may include elements like buttons, text, etc. in addition to natural images which enable performing a variety of tasks. Saliency in natural images is a well studied area. However, given the difference in what constitutes a mobile interface, and the usage context of these devices, we postulate that saliency prediction for mobile interface images requires a fresh approach. Mobile interface design involves operating on elements, the building blocks of the interface. We first collected eye-gaze data from mobile devices for free viewing task. Using this data, we develop a novel autoencoder based multi-scale deep learning model that provides saliency prediction at the mobile interface element level. Compared to saliency prediction approaches developed for natural images, we show that our approach performs significantly better on a range of established metrics.  

%We introduce models for saliency prediction for User Interfaces on mobile and small screen devices, and a novel dataset. Besides natural images, User Interfaces of mobile devices contain many salient regions or elements, like buttons, text, links, and natural images. There are often many visual cues which crowd the small screen. As a result, the viewing patterns of such interfaces may be different from natural images as well as web-based interfaces. We further introduce the concept of element-wise saliency which is necessary for images of User Interfaces. We propose new deep learning models which can predict saliency for such interfaces by learning User Interface specific features as well as incorporate features from natural images. Finally, we demonstrate the effectiveness of our models and setup for predicting saliency for the novel dataset.
\end{abstract}
%%%%%%%%% BODY TEXT
%!TEX root = main.tex

\section{Introduction}
\label{sec:introduction}

% Growth of mobile, speed of design
Mobile Devices have become ubiquitous in recent years and it has been accompanied by an explosion in the number of applications that are available for these devices. In the U.S. mobile apps overtook PC Internet on time spent in the year 2014~\cite{cnnmoney}. As the world moves towards pervasive mobile app usage, brands are increasingly trying to provide an engaging experience for their customers through them~\cite{cmo}. Developing apps constitutes a significant cost for brands~\cite{savvyapps}. One part of the app development process is designing applications likely to help the user performs tasks efficiently and in an engaging manner. The user interface (UI) design process today involves designers creating UI mocks, which are improved in an iterative manner. A part of the iterative process is the feedback from focus groups and peers. This is a time consuming and expensive process and presents opportunities for automation. We build models that can predict the saliency of different sections of a mobile app, and propose its use as a feedback tool for designers.

For desktop devices, eye-gaze tracking as a form of user engagement feedback has been studied~\cite{jacob2003eye}. Most desktop based eye-gaze tracking technologies rely on specialized hardware for capturing the human face and eyes while viewing \cite{saldatasets}. But such techniques cannot be applied to mobile device without compromising the natural usage pattern of such devices. However, modern mobile devices are almost always equipped with front facing cameras. Using these, it is possible to capture a user's face and eyes while she is exposed to a mobile screen. In this work, we leverage \textit{iTracker} \cite{krafka2016eye}, a Convolutional Neural Network (CNN) based model which can be used to predict the location of a users fixation on the screen, using the video feed from a front-facing camera as input.

\begin{comment}

\begin{figure}[t!]
  \centering
  \subfloat[]{\includegraphics[width=0.15\textwidth]{images/ui28.png}\label{fig:level1}}
  \subfloat[]{\includegraphics[width=0.15\textwidth]{images/pixui28.png}\label{fig:level2}}
  \subfloat[]{\includegraphics[width=0.15\textwidth]{images/eleui28.jpg}\label{fig:level0}}
  \caption{(a) Sample UI image, (b) Pixel-level saliency map, and (c) Element-level saliency map}
  \label{fig:sampleUI}
\end{figure}

\end{comment}

\begin{figure}[!t]
\centering
\includegraphics[scale=0.8]{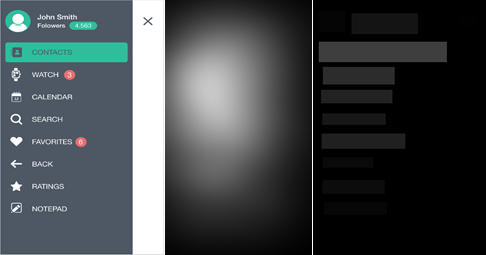}
\caption{A sample UI image, its pixel-level ground truth saliency map inferred from the collected front-facing camera video feed, and the element-level saliency map.}
\label{fig:sampleUI}
\end{figure}

One approach to predicting the saliency for mobile UIs is to predict pixel level saliency. But UI designers do not work with pixels, they work with \textit{elements}. We define a mobile UI element as the building blocks that are arranged to assemble the complete UI.  An element can be a natural image, button, text box or any other component present in the UI. During the design process, a designer can add, remove, edit, or change the relative position of an element. Given this fact, we decide to approach the problem as one of saliency detection at the element level. The saliency output must present a spatial coherence and a smooth transition between neighbouring pixels. Addressing eye gaze saliency at the element level preserves the spatial coherence and correlation for all the pixels of an element. It enables the designer to modify the design based on the relative saliencies of elements. Figure \ref{fig:sampleUI} shows a sample UI image with its collected pixel-level ground truth and element-level saliency maps.

%To address the problem of predicting the saliency for mobile UIs, we took an element-wise approach instead of a pixel wise saliency. An element can be an natural image, button, text or any other component present in the image of the interface. The motivation of doing this is because while looking at an interface, users tend to look at UI elements like buttons, icons, etc. rather than just looking at the individual pixels of the element. The saliency output must present a spatial coherence and a smooth transition between neighbouring pixels. Treating eye gaze saliency on element level preserves the spatial coherence and correlation for all the pixels of an element. Additionally, a UI designer always operates at the element level, either adding, removing, modifying or relocating an element. Thus, operating and optimizing our solution at the element level is the right approach. In Figure \ref{fig:sampleUI}, we have shown a sample UI image with its pixel-level and element-level saliency map.

In our work, we introduce models which can be trained on a dataset of UI images along with corresponding eye gaze data collected from users. This model can then be used to predict saliency for a new test UI, to provide rapid feedback to the designer. In summary, the main contributions of our work are as follows. We propose a novel model that uses de-noising autoencoders on multiple scales of UI elements to provide saliency prediction at the element level. For the task of saliency prediction in mobile UIs, we achieve accuracies which are significantly better than the state of the art in saliency prediction.\footnote{We can share our model with other researchers upon request.}
%\begin{itemize}
    %\item We create an eye gaze fixation dataset for $355$ mobile UI images, which is the first dataset on UI saliency, to the best of our knowledge.
 %   \item  We propose a nove model for UI saliency prediction which leverages multi-scale image features learned through a de-noising autoencoder.
 %   \item We propose element-level saliency computation for mobile UIs, this is in close alignment with how designers work.
%\end{itemize}

%The system makes predictions about where individuals are likely to fixate. Thus, we propose that the system will facilitate UI development during the iterative design process by removing the need for focus groups. Using this system, the designer can create and quickly compare variations of a UI design and understand which ones work better in terms of saliency. We will provide our model for element-level saliency prediction on request.

%This paper is organised as follows- In Section ?? we discuss related work, in Section ?? we describe the proposed model using . In Section ?? we report the quantitative and qualitative experimental results. Finally, we draw conclusions in Section ??.
%todo add applications

%!TEX root = main.tex

\section{Related Work}
\label{sec:relatedworks}

In this section, we summarize the four broad areas of research that have an implication on our work. 

\subsection{Saliency Models for Natural Images}
Predicting eye gaze for natural images is a well explored topic in computer vision. Some early natural image saliency methods were based on concepts like Feature Integration theory \cite{itti1998model}, graph-based normalization \cite{harel2007graph}, method that analyzes the log spectrum of image \cite{hou2012image}, information theory principles like self-information \cite{zhang2008sun}, 
%incremental coding length for defining spatio-temporal saliency \cite{hou2009dynamic},%
and information maximization \cite{bruce2006saliency, bruce2009saliency}.
Some models use supervised learning for saliency prediction based on manually designed feature sets  \cite{kienzle2007nonparametric, zhao2011learning, judd2009learning, borji2012boosting}. %Recent surveys \cite{borji2013state, li2014secrets} cover a lot of work in this field.
All these approaches modelled saliency in a bottom-up manner using low features, which leads to models that fail to generalize to complex scenes and new domains. 

Recent progress in saliency prediction has been driven by deep learning methods trained on large datasets which allows learning hierarchies of feature representations from the pixel level data. Some models like \cite{vig2014large} and SalNet \cite{pan2016shallow} have trained their own networks to predict saliency from scratch, while others have used features from pre-trained CNNs, such as DeepGaze \cite{kummerer2014deep}, SALICON \cite{huang2015salicon}, ML-NET \cite{mlnet2016}, and Deepfix \cite{kruthiventi2017deepfix}. More recent advances include methods like training using adversarial examples \cite{pan2017salgan} and neural attentive mechanisms to iteratively refine the predicted saliency map \cite{cornia2016predicting}. These methods are designed for natural image saliency prediction and we explore the applicability of these methods for mobile UI saliency prediction.

\subsection{Multi-Scale Feature Extraction}
Some recent models have attempted to explicitly model how the neighborhood of a location affects saliency at a particular location. Mr-CNN \cite{liu2015predicting} presents a multi-scale CNN which is trained from image regions centered on fixated and non-fixated image patches at multiple scales. A similar model is proposed in \cite{liu2016learning}. SALICON \cite{huang2015salicon} also incorporates features learned at two scales, coarse and fine, and optimizes KL divergence in the last layer. A multi-context approach over a subsampled and upsampled image patch at the super-pixel level has been proposed in \cite{zhao2015saliency}. Such methods try to leverage the contrast of an image region against the surrounding area for saliency prediction. All the methods mentioned so far have been developed exclusively for analyzing natural images, and are not trained or tested on graphic designs.

\subsection{Saliency and Attention Models for Webpages and Interfaces}
Attempts at understanding visual perception of interfaces and designs have been made since the last decade \cite{jacob2003eye}. One such work \cite{still2010saliency} predicts the entry point in webpages in a $50$ screen-shot dataset using features such as the center surrounded differences of colors, intensity, and orientations. A linear regression model on features extracted from HTML induced DOM to generate a model for predicting visual attention on webpages is explored in \cite{buscher2009you} (this work uses a dataset of $361$ webpages). In \cite{shen2014webpage}, a model combining multi-scale low-level feature responses, explicit face maps, and positional bias was proposed to predict fixations on the Webpage Image (FiWI) dataset, this dataset contains a total of $149$ screenshots of webpages. The work in \cite{shen2015predicting} extends this by replacing specific object detector with features from Deep Neural Networks. Users' mouse and keyboard input along with the UI components have been used in predicting their attention map \cite{xu2016spatio, predimportance}. A manually designed feature set to predict human visual attention for free-viewing webpages is studied in \cite{li2016webpage}. 

While this line of work presents the semantically closest area of research to our work, these are limited in their application only to webpages. Further given the size and structural differences of desktop webpages with mobile apps, these models cannot be directly ported to our problem. %Our work is the first to apply deep learning for User Interface saliency prediction in mobile UI interfaces at a coarser level, i.e. element-level while using just user's eye-gaze in natural settings as input.

%\subsection{Papers on Object Based Saliency}

\subsection{Eye-Gaze data Collection}
Traditionally, all work involving collection of eye gaze data has relied on custom hardware. For instance, all $23$ saliency datasets listed at \cite{saldatasets} are collected using custom hardware. The ubiquity of mobile devices pose unique challenges and opportunities. Some recent works have explored the possibility of using the front facing cameras of mobile devices to detect the eye gaze location of users looking at their mobile screen \cite{krafka2016eye, li2017towards}. Of these, we find \textit{iTracker} \cite{krafka2016eye}, a CNN based model, a more sophisticated approach. The \textit{iTracker} system has been developed for iOS, and we modify it to work on the Android OS based mobile phones. 

%!TEX root = main.tex

\section{Approach}
\label{sec:technical}
In this section, we describe the approach to saliency detection for mobile UIs. 

\subsection{Stimuli}
In the absence of any available eye-gaze datasets for mobile UIs, we created our own dataset by assembling a set of mobile UI images from $156$ android applications from the Google Play Store. We ensured that the selected apps represent a good spread with respect to their ratings (Table \ref{tab:ratings}) and download counts (Table \ref{tab:downloads}). For each application, $2$ UI screenshots were taken on an average, leading to a total of $293$ UI images. 
\begin{table}[ht]
\caption{Distribution of mobile apps for downloads}
\label{tab:downloads}
\begin{center}
 \begin{tabular}{| c | c | c | c | c |} 
 \hline
 \textless1 M  & 1 M & 5-10 M & 50-100 M & \textgreater100 M \\ [0.5ex] 
 \hline
 49 & 28 & 31 & 34 & 23 \\ 
 \hline
\end{tabular}
\end{center}
\caption{Distribution of mobile apps with ratings}
\label{tab:ratings}
\begin{center}
 \begin{tabular}{| c | c | c |} 
 \hline
 2-3 & 3-4 & 4-5 \\ [0.5ex] 
 \hline
 6 & 33 & 114 \\ 
 \hline
\end{tabular}
\end{center}
\end{table}

\begin{figure}[t!]
\centering
\includegraphics[scale=0.4]{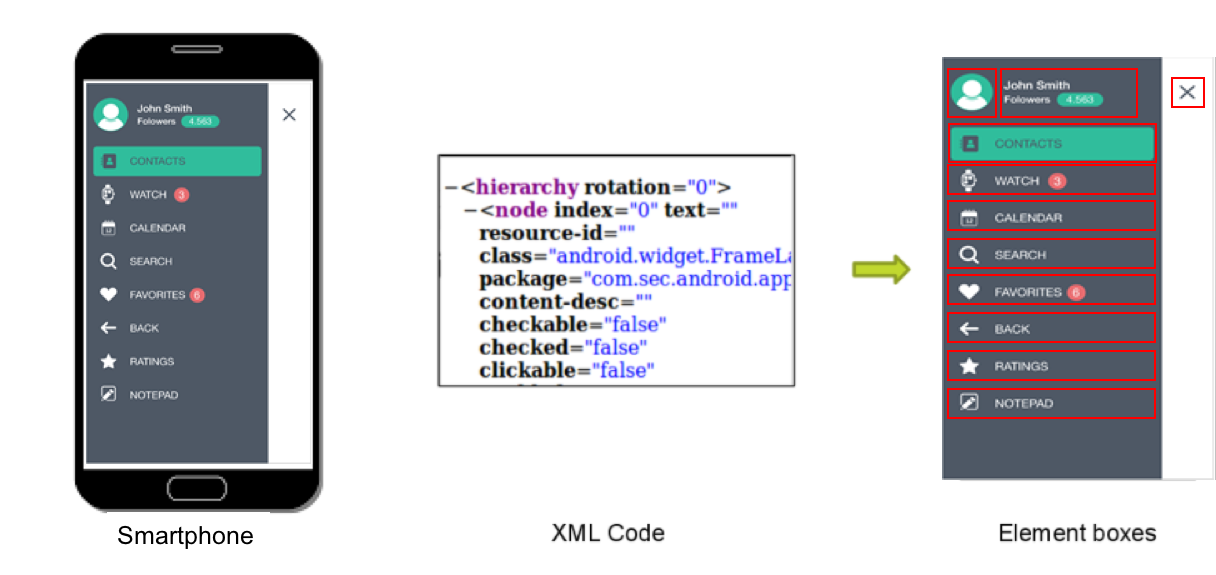}
\caption{Element box extraction}
\label{fig:elementapp}
\end{figure}

\begin{figure}[t!]
\centering
\includegraphics[scale=0.5]{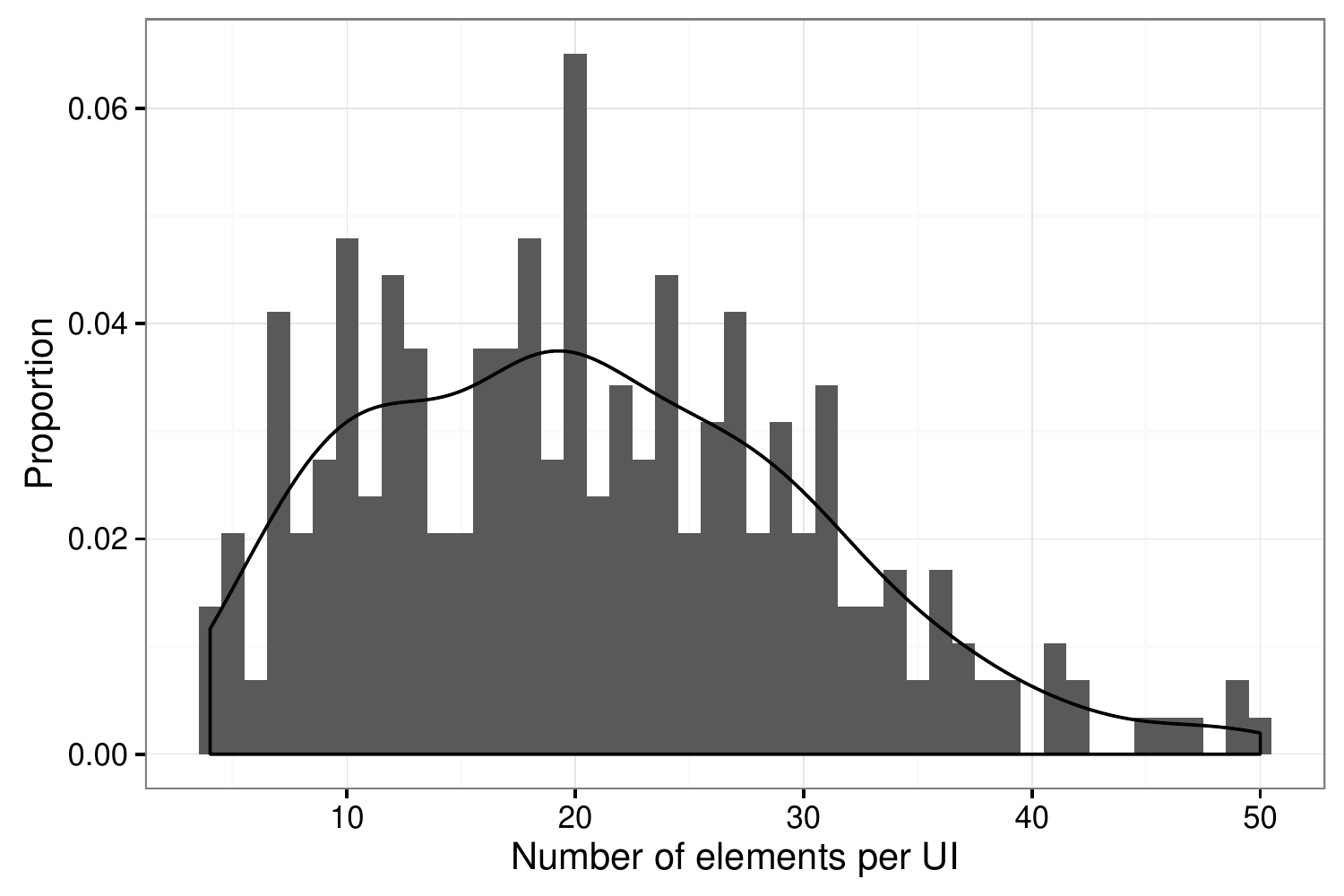}
\caption{Histogram of elements per image, the curve represents the density estimator.}
\label{fig:numelement}
\end{figure}

Since our goal was to predict saliency at an element level, the bounding boxes of the elements were required. For this, two methods were used. In the first method, while capturing screenshots for the mobile UIs, we process the logs from the official Android debug tools to get information about the underlying XML code of the application. The XML code was processed to obtain bounding boxes of elements present in the UI. While doing this, a smaller element by area was considered to be 'over' a larger element so that a pixel belonging to more than one element is assigned the ID of smaller element. This method does not work in scenarios where UI elements have a lot of overlaps, and so we use another method which involved semi-automated a drag and drop scheme to generate the bounding boxes. %This required manually selecting the elements by using drag and drop across its main diagonal, considering them as quadrilaterals. %, or just a click on the element in case the appropriate bounding is detected.
One example output is shown in Figure \ref{fig:elementapp}. The distribution of the number of elements per UI is shown in Figure \ref{fig:numelement}. The mean number of elements per UI is $20.64$ (Standard Deviation of $9.80$), with all UIs having at least $4$ and at most $50$ elements. 

\subsection{Eye-gaze Data Collection Experiment}
In order to collect free viewing eye-gaze data for the set of mobile UI images described in the previous section, we developed a mobile application which displays screenshots of mobile UI images to participants in a natural environment.
%We carried out the free-viewing experiment in two phases. In the first phase, $30$ participants used the same mobile device ($1080 \times 1920$ resolution) in a consistent environment. The participants belonged to the age group of $19-26$ with $4$ female participants. Proper lighting condition was ensured. The second phase was carried out on Mechanical Turk, in which 80 people participated. Participants were given comprehensive instructions on how to download and use the application and participate in the experiment. %
We conducted an experiment on Mechanical Turk, where $111$ participants downloaded our application on their mobile devices for the experiment. The participants belonged to the age group of $19-46$. Participants were given comprehensive instructions on how to download and use the application to participate in the experiment. The application collected the front facing camera's video feed from each participant across multiple sessions and this feed was sent to our server. 

Each session began with a calibration task (described in section \ref{sec:procgaze}) which was followed by displaying 10 mobile UI images for free viewing, that is, no instructions were given to perform specific tasks. The mobile interface screenshots were interspersed with filler images with a probability of occurrence of $0.33$. This was done to remove any spatial bias from from prior images. The filler images consisted of sceneries and abstract art, no video feed was collected for these images. Each participant was shown an average of $50$ different UI images across a span of $7$ or $8$ sessions. Each UI image persisted for $5$ seconds with a $3$ second gap in between each image. The participants were free to pause between sessions in case they wanted to take a break. In the experiment, each UI screenshot was shown to an average of $9$ different participants, while ensuring that the same image is not shown twice to the same participant.

\subsection{Processing Eye-gaze Data}
\label{sec:procgaze}
From the videos of the participants collected for the free viewing task from the previous step, we generated the gaze points which correspond to where the participants were looking at in the various UI images that were shown to them. To achieve this we started with \textit{iTracker} \cite{krafka2016eye}. This work introduced a eye tracking software that works on devices such as mobile phones and tablets, without the need for additional sensors or devices. While the available software is designed for iOS devices, we modified it to run on Android devices. The captured videos were split into frames. For each frame, the crops of face and both eyes were generated using \cite{dlib09}. These are required as input for \textit{iTracker}. The output is the $(x,y)$ coordinates of the gaze point corresponding to each frame of the video. We can use this to locate the pixel of the UI a participant is looking at in each frame.
% The architecture and sample input and output of \textit{iTracker} are shown in Figure \ref{fig:iTracker}. 

% \begin{figure}[h!]
% \centering
% \includegraphics[scale=0.30]{images/itracker1.jpg}
% \caption{iTracker}
% \label{fig:iTracker}
% \end{figure}

The prediction of the gaze points from \textit{iTracker} was found not accurate enough for the task at hand (with an average error from $3.5$ cm). As a solution to this, we included a calibration task at the beginning of each session of the app. During this task, we showed a moving object at $11$ different positions on the screen for a total of $20$ seconds. The participants were instructed to follow the object on the screen. The video of the participant captured during the calibration task was processed as described earlier. 

A linear regression model was trained for the calibration sections of each session with the gaze points predicted by \textit{iTracker} as the features, to predict the actual coordinates of the object shown. We divided the calibration frames into training and test sets in a $3:1$ ratio and measured the tracking error on the test set. The calibration task helped in reducing the average error from $3.5$ cm to $1.4$ cm with mean standard deviation of $0.89$ cm. This error is in the range of error reported in paper~\cite{krafka2016eye}. We used the regression output as the gaze point and also generated a 2-dimensional co-variance matrix. This is utilized during processing of ground truth eye-gaze fixations. 

\subsection{Generating Saliency Maps from Fixations}

\subsubsection{Pixel-level Saliency}
We use the eye fixations predicted using the calibrated \textit{iTracker} outputs for calculating the probability of a fixation point falling on a pixel. The 2-dimensional co-variance matrix generated during calibration was used in Gaussian blurring of the fixation maps for each UI viewed by the participant for a session. Through this procedure, we get a pixel-level probabilistic heatmap from the fixation points. Converting fixation locations to a continuous distribution allows for uncertainty in the eye-tracking ground truth measurements to be incorporated, as suggested in \cite{bylinskii2016different}. We leverage the error from calibration as it varied from one session to another based on how the mobile was held and the lighting conditions. 

\begin{figure}[!t]
\centering
\includegraphics[scale=0.06]{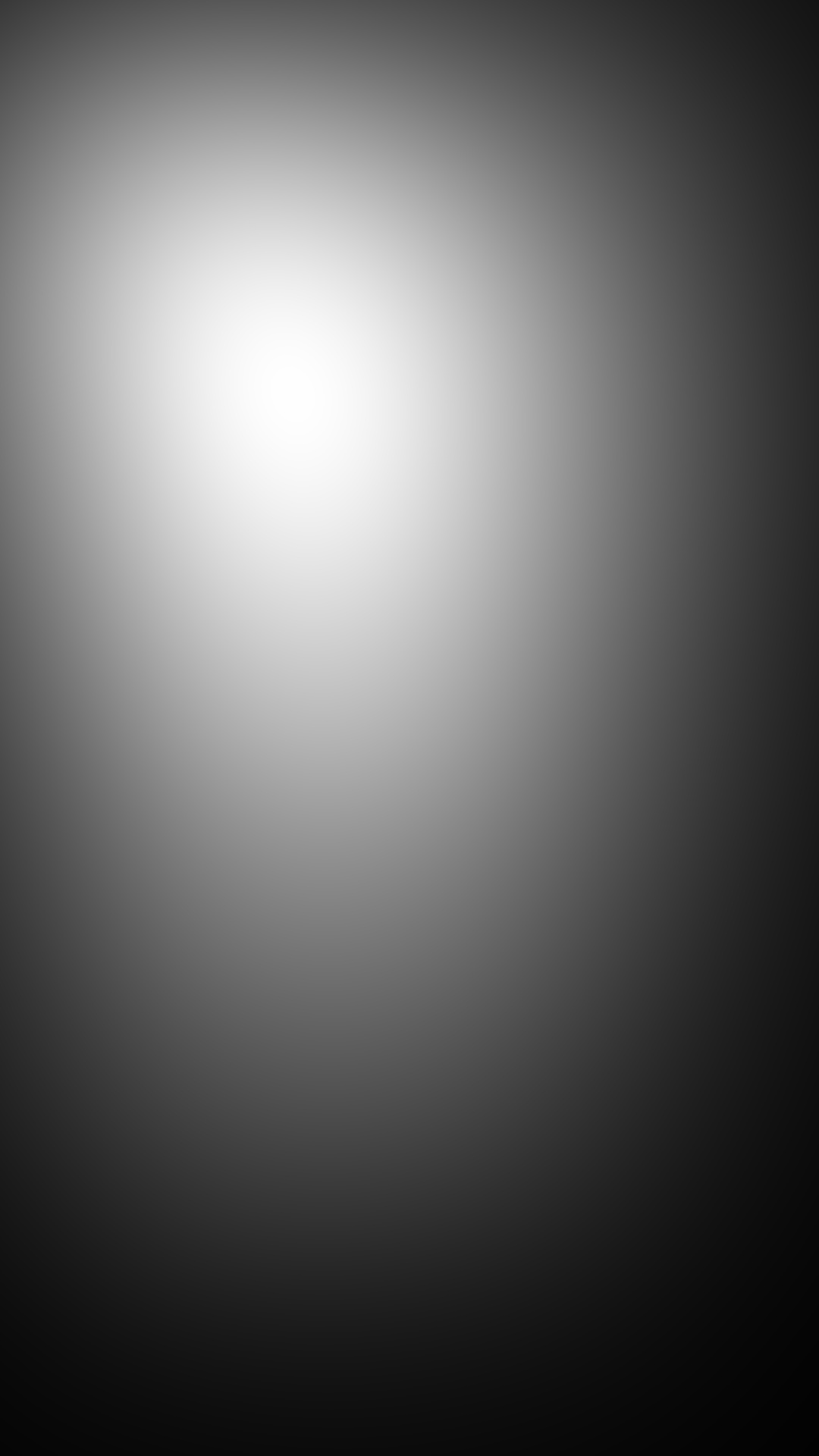}
\caption{Average ground truth saliency map}
\label{fig:alluiaverage}
\end{figure}

The average saliency map can be seen in Figure \ref{fig:alluiaverage}. It indicates a top-left bias similar to the webpage saliency dataset in \cite{shen2015predicting}. This is primarily because important UI elements are generally present in this area, and participants tend to browse the images top-to-bottom and left-to-right.

\subsubsection{Element-level Saliency}
We convert the pixel-level saliency maps into the UI element-level saliency maps. For this, we compute the integral of the pixel-level saliency density over the area covered by an element. This is followed by normalization over all elements to ensure that the vector sums to $1$. Given an UI with $k$ elements, we represent the element saliency map $\mathbf{E} = (E_1, \cdots, E_k)'$, as vector of probabilities where $E_j$ is the probability of element $j$ being fixated. In case one UI element overlaps another element, we assign the saliency of the pixels in overlapping regions to the element on top. Sample pixel-level and element-level saliency maps are presented in Figure \ref{fig:qualitative}.

\subsection{Feature Extraction from UI Images}
Saliency is driven by visual contrast and it indicates which parts of an image are more visually appealing relative to the rest of the image. Thus, the saliency model needs to capture the contrast between a region of the image, a UI element in our case, and its surrounding area. Therefore, we extract features for every UI element at three scales. The first scale is the image of the UI element itself. The second scale consists of the UI element along with a region surrounding it, whose boundaries are decided by the mid-point of the element's boundary and the entire UI image’s boundary for both dimensions. The third scale consists of the entire UI image. Our saliency models, described in detail in later sections, then uses these multi-scale features along with other low level features to train fully connected neural network layers for saliency prediction at an UI element-level. We now describe our feature generation methods.

\subsubsection{Feature Extraction from Stacked De-noising Autoencoder}
\label{sec:autoencoder}
We use an autoencoder model for learning feature representation for our saliency models as they provide an effective way to learn good feature representations by using large amount of unlabeled data \cite{vincent2010stacked}. Autoencoders are neural networks that consist of two parts, an encoder and a decoder. The encoder reduces the input to a lower dimensional representation and the decoder reconstructs it into the original input. The objective of the autoencoder is to enforce an output to be as close as possible to the corresponding input. 

However, it is proven that the reconstruction criterion is not enough in itself to guarantee the extraction of useful features, as it suffers from non-generalizability. A good feature representation should be stable and robust under corruptions of the input and should capture useful structure in the input distribution. It has been shown that feature extractors learned by de-noising autoencoders are able to learn useful structure in the data, that regular autoencoders seemed unable to learn \cite{vincent2010stacked}. We adopt the concept of de-noising autoencoders to learn such a representation for images, where the input image is corrupted by setting a fraction $f$ of the pixels of the image to $0$. Let's call this noisy version of the image $i$, $\tilde{x}_i$. The de-noising autoencoder tries to reconstruct the original image $x_i$ by producing reconstruction $\hat{x}_i$. It minimizes the reconstruction error using the Euclidean loss,

$$  \min L =  \sum_{i = 1}^K ||\hat{x}_i - x_i||^2.$$

The architecture of our autoencoder consists of $5$ convolutional layers. We adopt $3$ and $16$ filters with size $(3\times3)$ and a stride of $1$ in the first two convolutional layers, respectively. Both are succeeded  by Max-pooling layers. All max-pooling layers have size of  $(3\times)3$ and a stride of 1. This is the encoder part of our autoencoder. this is followed by another three convolutional layers with size $(3\times3)$ and a stride of 1 and with $16$, $32$ and $3$ filters, respectively. After the the third and forth convolutional layers we add upsampling layers with size of $3\times3$. All convolutional layers use ReLU activations \cite{nair2010rectified}. The encoder part of the autoencoder converted a $(288\times162\times3)$ size input image into a $(32\times18\times16)$ sized encoded output.

\begin{figure}[t]
    \centering
  \includegraphics[width=0.18\textwidth]{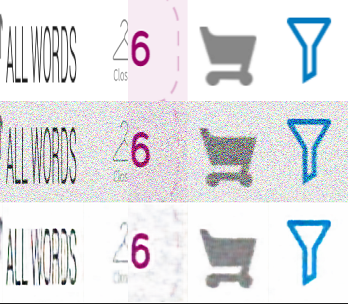} 
  \caption{From top to bottom - sample images of elements, their generated noisy versions, and their reconstructed versions. Only the first of three scales is displayed.}
  \label{fig:autosamples}
\end{figure}

We trained the autoencoder on all UI images in our dataset. By using autoencoders we were able to learn features for the UI images and at the same time reduce the input dimensions needed for the saliency prediction model. We learned $3$ separate autoencoder models for the $3$ scales of the elements independently. All three autoencoders share the same architecture but have different parameter values. In Section \ref{sec:saliencymodel} we will talk in more detail about how the autoencoder model contributes to the saliency prediction. Some sample UI elements, their noisy versions and their reconstructed versions are shown in Figure \ref{fig:autosamples}. 

% \begin{figure}[h!]
% \centering
% \includegraphics[scale=0.23]{images/level0.png}
% \caption{level 0}
% \label{fig:auto0}
% \end{figure}
% \begin{figure}[h!]
% \centering
% \includegraphics[scale=0.23]{images/level1.png}
% \caption{level 1}
% \label{fig:auto1}
% \end{figure}\begin{figure}[h!]
% \centering
% \includegraphics[scale=0.43]{images/level2.png}
% \caption{level 2}
% \label{fig:auto2}
% \end{figure}

\begin{comment}
\begin{figure}[h!]
  \centering
  
  \subfloat[Scale 2]{\includegraphics[width=0.23\textwidth]{images/level2.png}\label{fig:level1}}
  \hfill
  \subfloat[Scale 1]{\includegraphics[width=0.23\textwidth]{images/level1.png}\label{fig:level2}}
  \vfill
  \subfloat[Scale 0]{\includegraphics[width=0.18\textwidth]{images/level0.png}\label{fig:level0}}
  \caption{In each image, from top to bottom- Sample images of elements at a scale, their noisy versions used for testing and their reconstructed versions.}
  \label{fig:autosamples}
\end{figure}
\end{comment}

\begin{figure*}[th!]
    \centering
  \includegraphics[scale=0.5]{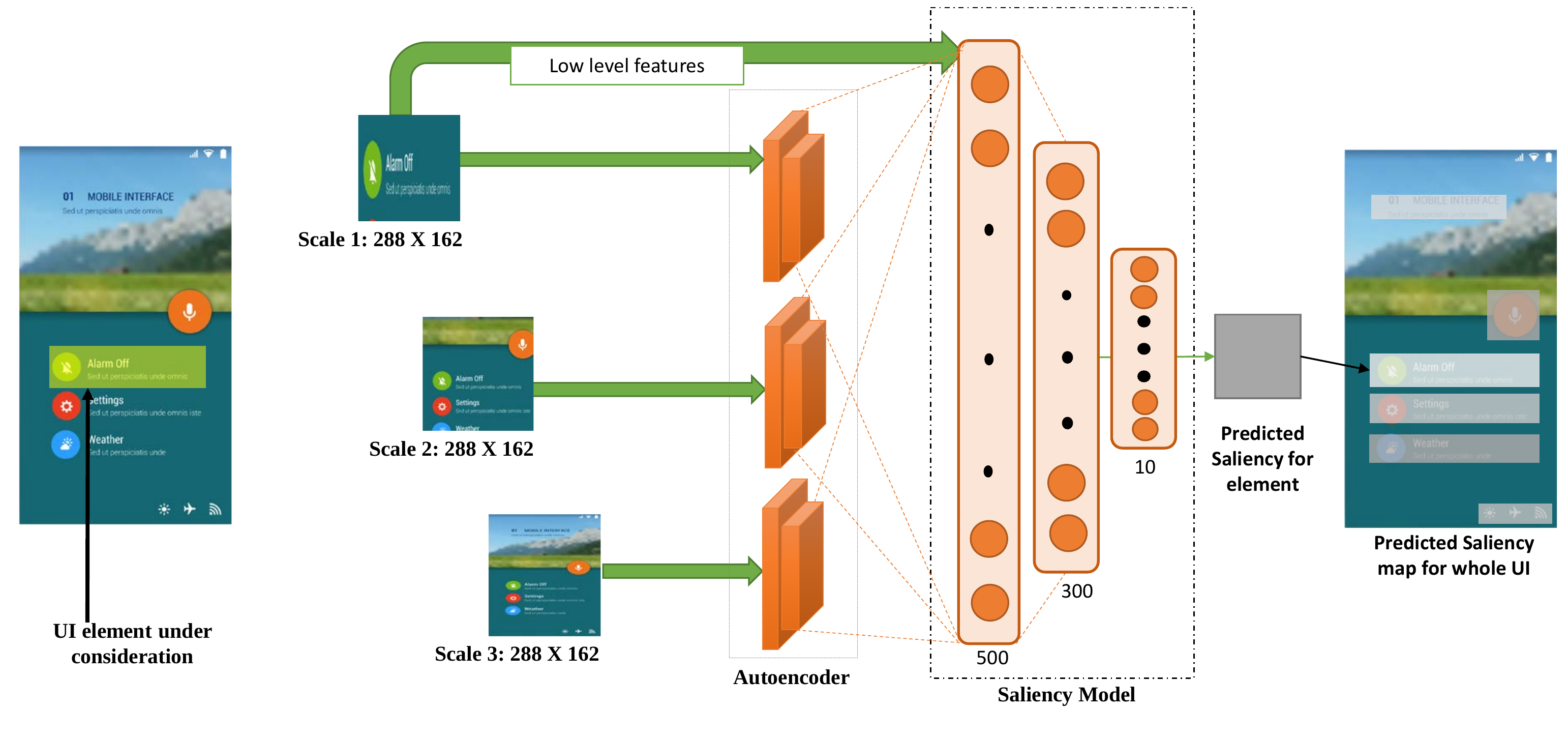}
  \caption{The overall architecture of the mobile user interface saliency prediction system. For each UI, we first segment it into elements. In the above example, we predict the saliency of the ``Alarm off" element. In addition this this element, we also take two high zoomed out images of the element we call scales. The autoencoder versions of all three are fed into the deep model. The saliency of all the elements on the page is reconstructed by combining the saliency of individual elements. }
  \label{fig:architecture}
\end{figure*}

\subsubsection{Low-level Feature Extraction}
\label{sec:lowlevel}
We also computed low-level features based on color distribution, size and position of the elements in the UI image. There were $16$ such features generated for each element in the UI, including width, height, area and position in pixel coordinates, along with the first and second color moment for each color channel \cite{yu2002color} of both the element under consideration, as well as the whole image separately. We have included area, width and height in the feature set since we are rescaling all the elements before they are input to the model, and thus we loose the information regarding their size and scale in the process. The features for position helps in capturing the user's bias towards UI elements at the top and left of the screen. For $N$ pixels in the image or element and $p_{ij}$ as value of the $j^{th}$ pixel of the image at the $i^{th}$ color channel, the first color moment $M_i$, analogous to the mean; and the second color moment $\sigma_i$, analogous to the standard deviation, can be calculated by
\begin{align*}
  M_i =
  \sum_{j = 1}^N \frac{1}{N}p_{ij}, \text{ and }  \sigma_i =
  \sqrt{\frac{1}{N}\sum_{j = 1}^N(p_{ij}-M_i)^2}. 
\end{align*}

\begin{comment}
\subsubsection{Features from SalNet}
In one of our models, we include saliency related features from SalNet \cite{pan2016shallow}, which is the state-of-the-art in saliency prediction on natural images. It is a deep neural network that was trained on a large collection of natural images. However, SalNet has not been trained on UI images. Additionally the output generated by SalNet is at a pixel-level and not at an element-level like we want to generate. Their network leverages the VGG network's \cite{simonyan2014very} lower layers trained for classification, for the task of saliency prediction. We fed UI images as input to the pretrained SalNet and extracted features from its second-last convolution layer with dimensions of ($32\times80\times45$). We then applied PCA on this 32 dimensional to reduce it to 10 across all UI images. From this, we get 20 features per UI while retaining 95\% of the variance in PCA.
\end{comment}

\subsection{Saliency Prediction Model}
\label{sec:saliencymodel}
Our primary aim is to predict the eye-gaze fixations at an element level. For each element, we predict the probability of fixation on the element by incorporating features learned at the three scales of the UI element and some low-level features. Motivated by works such as \cite{liu2015predicting, liu2016learning}, we combine information from the three scales to incorporate both the local and the global contrasts to infer the saliency. Combining features at different levels has been shown to increase the performance of predicting the saliency map \cite{judd2009learning, xu2014predicting}. The idea behind this is that the saliency of an element depends not only on the element itself, but also on the content surrounding the element. 

%Our model incorporates features learned from raw pixel data using an autoencoder and a set of low-level features, as shown in Figure \ref{fig:architecture}. We call this model $\mu$-Nc.

The architecture of our model is shown in Figure \ref{fig:architecture}. For each element, we generated crops of the element from the UI at $3$ scales, as described earlier. For image regions at each scale, we first resize them to the size of $288\times162$ disregarding their aspect ratio. This is done so that the autoencoder models at each resolution level can share the same architecture. Then, the features coming from different scales are fed into the three convolutional streams of the autoencoder. The details of the autoencoder model are mentioned in the section \ref{sec:autoencoder}. The output of the three parallel streams is concatenated with the low-level features mention in section \ref{sec:lowlevel} and becomes the input for the subsequent three fully connected layers. These layers learn to predict saliency of the element with respect to its appearance as well its neighborhood. We used the ReLU \cite{nair2010rectified} activation in all layers due to its superior effectiveness and efficiency. Dropout layers were used in between every pair of fully connected layers in order to prevent over-fitting as suggested in \cite{hinton2012improving}.

The element-level ground truth saliency maps are normalized in the range of $(0,1)$. But, since each UI has different number of elements, we do not have a response of a consistent dimension. Hence, we treat prediction for each element as independent of the others. We apply an element-wise activation function in the final layer, and treat the element-wise predictions as probabilities for independent binary random variables. We can then apply the binary cross entropy (BCE) loss function $L(E_i, \hat{E_i})$ between the predicted element-wise saliency map $\hat{E}$ and corresponding ground truth $E$ in this setting. We also experimented with mean squared error (MSE) or Euclidean loss which has been successfully applied in similar settings \cite{mlnet2016, pan2016shallow}, but we found that BCE performed better in the experiments. 
As described earlier, a number of saliency approaches for natural images has been studied in the literature. We hypothesized that leveraging the knowledge contained in these models may provide valuable information to our model. To this end, we proposed another model called $\mu$-SalNc. This model uses features from SalNet \cite{pan2016shallow}. We generate a feature vector of dimension $(80\times45)$ by providing the third level scale for each UI element through SalNet's penultimate convolutional layer. This vector is concatenated with the features from the autoencoder and low level features, a learning performed through a fully connected and dropout layers, similar to the $\mu$-Nc model.

\if
The $3$ scales of the images generated for each UI element was passed through SalNet independently and the features from the last layer of dimension $(80\times45)$ was extracted for each of them. The features extracted for all the $3$ levels were stacked depth-wise to generate a feature vector of dimension of size $(3\times80\times45)$ as an input to our model \textit{$\mu$-Sal}. This model consisted of a convolutional and max-pooling layer at beginning followed by a flatten layer and a network of fully connected and dropout layers. Max-Pooling layers were used to reduce the size and it was flattened to convert it into a single dimensional vector. Dropout layers were used in between every pair of dense layers in order to prevent over-fitting.
\fi

\section{Experiments and Results}

\subsection{Training and Validation on Mobile UI Dataset}
We trained and validated our model on our dataset using 4-fold cross validation, which consists of eye-gaze of users on $293$ Mobile UI screenshot images. For generating saliency maps for each UI element in the test image set, first saliency of each element is predicted. Predicted saliency value of all elements in the test image is then normalized so that the total saliency is $1$. This is done since the saliency of UI elements in the UI image is a probability distribution (positive numbers adding to 1 for a UI). The network was trained using stochastic gradient descent with a Euclidean loss. We used a batch size of $30$. The network was validated against a validation set after every iterations to monitor convergence and over-fitting. We used the standard $L2$ weight regularizer and ADAM optimizer \cite{kingma2014adam}. The autoencoder took approximately 15 hours and the saliency model took approximately 6 hours to train for 1000 epochs on a machine with 4 NVIDIA K520 GPUs.

\subsection{Evaluation Metrics}
We evaluate our approach on using three metrics. We describe these next. Our approach makes a saliency prediction for all elements that comprise a UI. The evaluation metrics thus apply on the vector of element level saliencies, the ground truth and the predictions. Denote the vector of ground truth as $\mathbf{E} = (E_1, E_2, \cdots, E_k)'$. Further, let the predicted saliencies be $\hat{\mathbf{E}} = (\hat{E_1}, \hat{E_2}, \cdots, \hat{E_k})'$. 

\textbf{AUC} or area under the Receiver Operating Characteristics (ROC) curve is the most widely used score for saliency model evaluation. In AUC computation, the estimated saliency map is used as a binary classifier to separate the positive samples (human fixations) from the negatives (the rest). By varying the threshold on the saliency map, an ROC curve can then be plotted as the true positive rate vs. false negative rate. AUC is then calculated as the area under this curve. For the AUC score, 1 means perfect prediction while 0.5 indicates chance level. 
\if
However, AUC can easily be influenced by center-bias, which makes a fair model comparison difficult.
\fi
AUC requires discrete fixations in its calculation. We chose the top 20 percent salient UI elements of the image to form the ground truth continuous saliency map as actual fixations, similar to what is described in \cite{judd2009learning}. We report the average AUC for all test images. 

\textbf{CC} measures the linear correlation between the estimated saliency map and the ground truth fixation map, i.e., the correlation between vectors $\mathbf{E}$ and $\hat{\mathbf{E}}$. These are then averaged over all the UI images in the test set. The closer CC is to 1, the better the performance of the saliency algorithm. 

\textbf{KL} divergence is a measure of distance that captures the distance between a target and predicted distribution. It assigns a lower score to a better approximation of the ground truth by the saliency map. All  metrics  have their advantages and limitations and a model that performs well should have relatively high score in all these metrics.

\begin{table}[t!]
\centering
\caption{Comparison of proposed approaches, against the baselines}
\label{tab:results}
\begin{tabular}{|llll|}
\hline 
Method 	 	  & AUC $\uparrow$	 	 	& CC $\uparrow$ 				& KL $\downarrow$		 	  	\\ \hline \hline
$\mu$-Nc 	  & \textbf{0.9256}& \textbf{0.8197}&      	\textbf{0.2340} 	 	 \\ 
$\mu$-SalNc &  0.9212 	 	& 0.8094        & 0.2882 	 	  \\ \hline \hline
GBVS 	 	  & 0.8751 	 	& 0.7613  		& 0.2465 	 	 \\ 
Itti 	 	  & 0.8423 	 	& 0.7019   		& 0.2843 	 	 \\ 
SalNet 	 	  & 0.8725 	 	& 0.7671  		& 0.2495 	\\ 
SalGAN 	 	  & 0.8482 	 	& 0.6894 		& 0.3318 	 	 \\ 
SAM 	 	  & 0.7316 	 	& 0.4927  		& 1.2603 	 	 \\ 
ML-NET 	 	  & 0.8703 	 	& 0.7541  		& 0.2678 	 	 \\ 
OPENSALICON &  0.8629 	 	& 0.7358  		& 0.2694 	 	 \\
Lab-Signature & 0.7971 	 	& 0.5177  		& 0.4368 	 	 \\ \hline 
\end{tabular}
\end{table}

\begin{figure*}[t!]
  \includegraphics[width=17.5cm, height= 12.5cm]{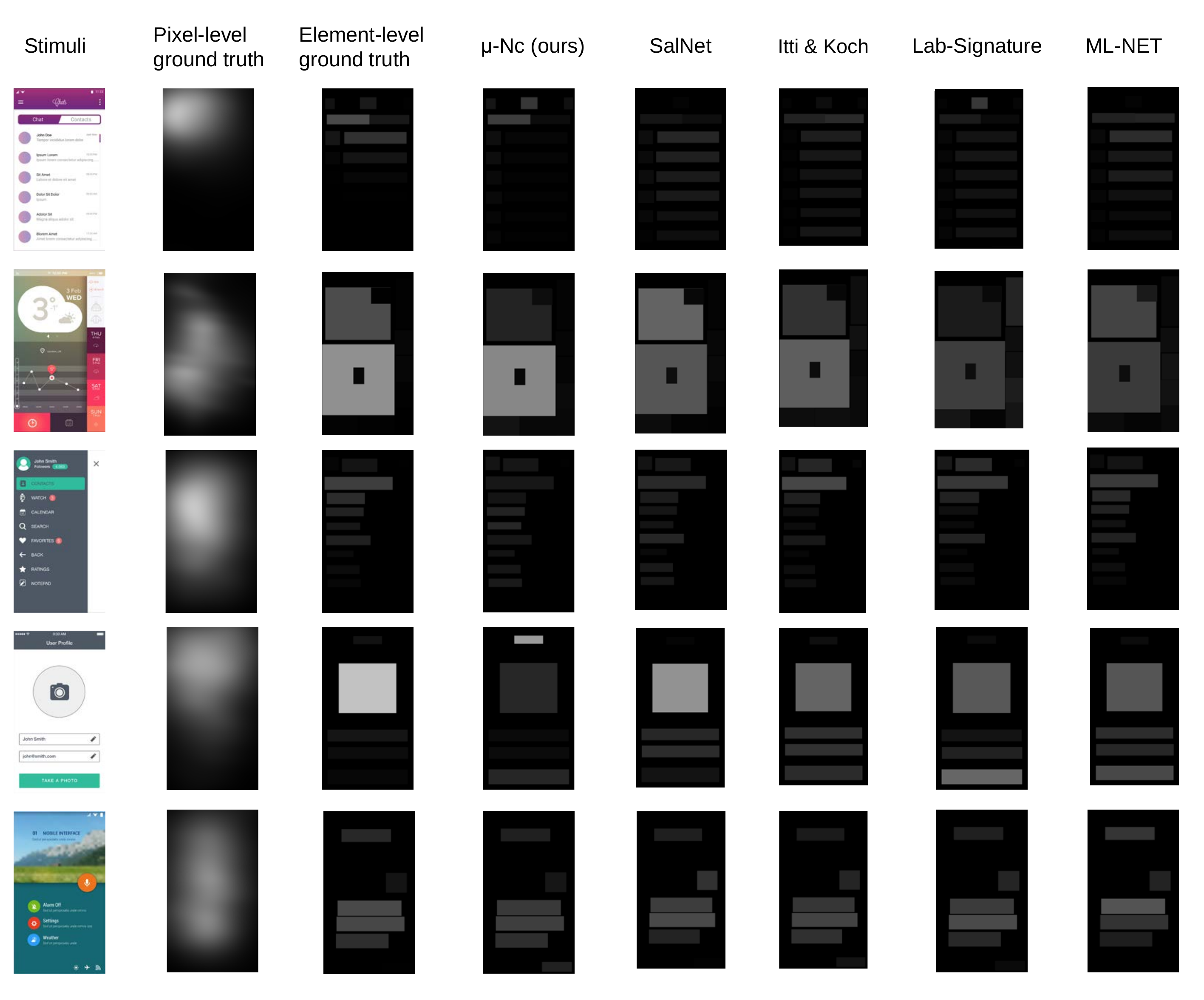}
  \caption{Qualitative comparison of our model with other methods}
  \label{fig:qualitative}
\end{figure*}

\subsection{Baselines and Comparisons}
We compare our work with GBVS \cite{harel2007graph}, Itti \cite{itti1998model}, SalNet \cite{pan2016shallow}, SalGAN \cite{pan2017salgan}, SAM \cite{cornia2016predicting}, ML-NET \cite{mlnet2016}, OPENSALICON \cite{huang2015salicon}, and Lab-Signature \cite{hou2012image}. We use the pretrained models from the prior work, predicting saliency for resized images. All these models provide pixel-level saliency distributions, which are then converted to the element level saliency distributions (following the same procedure used for the ground truth) by summing over all pixels belonging to an element and dividing by the sum across all elements. For our model, we carried out 4-fold cross validation for reporting the results. We didn't compare with prior work related to webpage saliency detection because webpages have very different size, number of elements, and orientation. Further, there were no open implementations available for comparison.

\subsection{Discussions}
The results on the $3$ evaluation metrics mentioned in the previous sections are presented in Table \ref{tab:results}. On all three metrics, we observe that $\mu$-Nc performs the best. This approach is better than $\mu$-SalNc. In other words, including features from saliency models for natural images, like SalNet, in the training model does not provide a better model for learning saliency in mobile UIs. Compared to the next best natural image saliency models, the proposed approach is 6\%, 7\%, and 5\% better on AUC, CC, and KL metrics, respectively. This shows that the best models trained for predicting saliency for natural images falls short for mobile UIs. Thus justifying the need to address this problem anew. The proposed approach has two advantages over existing saliency prediction models. First, by collecting eye-gaze data for mobile UIs, we inform our model of patterns which are unique to mobile UIs and their viewing in natural environments. Second, by modeling saliency at the element level, we optimize our model to predict element level saliency, leading to superior predictions.
%On AUC, we observe that both $\mu$-Nc and $\mu$-SalNc are better than other approaches, by a margin of $0.04$. With $\mu$-Nc slightly outperforming $\mu$-SalNc. This shows that including features from a model trained for natural image saliency is not suitable for saliency prediction on UI images. Since we compute the AUC for the top 20\% most salient elements within a UI, it shows that both proposed methods are better at detecting the most salient elements within a UI. On CC, $\mu$-Nc and $\mu$-SalNc perform better than all other methods, with the latter doing slightly better. Finally, on KL (lower being better), GBVS performs the best followed by SalNet, MLNET and OPENSALICON. The differing performance of the proposed methods on AUC and CC compared to KL arises from the difference in what these metrics are measuring. The KL divergence is a measure of distance that captures the entire distribution, where as AUC captures the head of the distribution and CC captures linearity in the comparison. These results mean that the proposed approach performs the best in identifying high saliency elements, and detecting the overall trend, but fails to capture the entire distribution as well as the other methods. 

Figure \ref{fig:qualitative} presents a qualitative analysis of the different approaches' performance in element level saliency prediction. For $5$ sample UI images, we show the original image, pixel-level ground truth, element-level ground truth, predictions form $\mu$-Nc, \cite{pan2016shallow}, \cite{itti1998model}, \cite{hou2012image} and \cite{mlnet2016} in the columns. Brighter shades indicate more saliency. Here are some observations - First, the ground truth reflects a top left bias for most images, and this is also reflected in the predictions from the different models. Second, other models tend to favour larger elements, where as $\mu$-Nc sometimes predicts small elements to have high saliency. TSalNet and $\mu$-Nc often predicts a skewed distribution, with some elements predicted to have high and some have low saliencies. The other approaches tend to predict a flatter distribution. Fourth, our model has learnt to give higher saliency to some UI specific components like text, specially if it has high contrast with the background. Also, instead of just simply relating saliency with low-level features such as color contrast, our model also incorporates the surrounding region in consideration while making predictions as can be seen in the prediction for the mic element for the fifth sample UI.
\
\section{Conclusions and Future work}
This paper presents a novel deep learning architecture for saliency prediction on mobile UI images. Our model learns a non-linear combination of low and high level features to predict saliency at an the element level. Qualitative and quantitative comparisons with state of the art approaches demonstrate the effectiveness of the proposed model. Learning from eye-gaze data on mobile UIs and predicting at the element level leads to a more accurate saliency model for mobile UIs. Our proposed model of element level saliency predictions can help a UI designer make decisions in the following manner. The designer can make changes to properties like the color, size, and aspect ratio at the element level; the number of elements and relative positioning at the UI level. For each modification, the designer can receive feedback on these changes in terms of saliency. The designer can also use this to compare and decide among a set of variants of the same UI.

In future work, we will explore a model which learns to simultaneously detect UI components and predict saliency. Another direction of research is to understand the ease of task completion for mobile UI through eye-gaze patterns. 

\begin{comment}
\begin{table}[t!]
\centering
\caption{Comparison of proposed approaches, against the baselines}
\label{tab:results}
\begin{tabular}{|lllll|}
\hline 
Method & AUC & KL & CC & MSE \\ \hline \hline
$\mu$-Nc & \textbf{0.9323} & 0.3540 & 0.7953 & 0.0032      \\ 
$\mu$-SalNc & 0.9212 &0.2882 & \textbf{0.8094} & 0.0035    \\ \hline \hline
GBVS & 0.8751 & 0.2465 & 0.7613 & \textbf{0.0028} \\ 
Itti & 0.8423 & 0.2843 & 0.7019 & 0.0030  \\ 
SalNet & 0.8725 & \textbf{0.2495} & 0.7671 & 0.0028 \\ 
SalGAN & 0.8482 & 0.3318 & 0.6894 & 0.0041\\ 
SAM & 0.7316 & 1.2603 & 0.4927 & 0.0067 \\ 
ML-NET & 0.8703 & 0.2678 & 0.7541 & 0.0030 \\ 
OPENSALICON & 0.8629 & 0.2694 & 0.7358 & 0.0032 \\
Lab-Signature & 0.7971 & 0.4368 & 0.5177 & 0.0060 \\ \hline 
\end{tabular}
\end{table}
\end{comment}
%-------------------------------------------------------------------------

\bibliographystyle{ieee}
\bibliography{references}

\begin{thebibliography}{10}\itemsep=-1pt

\bibitem{saldatasets}
{MIT Saliency Benchmark}.
\newblock \url{http://saliency.mit.edu/datasets.html}, 2017.
\newblock [Online; accessed 09-September-2017].

\bibitem{borji2012boosting}
A.~Borji.
\newblock Boosting bottom-up and top-down visual features for saliency
  estimation.
\newblock In {\em Computer Vision and Pattern Recognition (CVPR), 2012 IEEE
  Conference on}, pages 438--445. IEEE, 2012.

\bibitem{bruce2006saliency}
N.~Bruce and J.~Tsotsos.
\newblock Saliency based on information maximization.
\newblock In {\em Advances in neural information processing systems}, pages
  155--162, 2006.

\bibitem{bruce2009saliency}
N.~D. Bruce and J.~K. Tsotsos.
\newblock Saliency, attention, and visual search: An information theoretic
  approach.
\newblock {\em Journal of vision}, 9(3):5--5, 2009.

\bibitem{buscher2009you}
G.~Buscher, E.~Cutrell, and M.~R. Morris.
\newblock What do you see when you're surfing?: using eye tracking to predict
  salient regions of web pages.
\newblock In {\em Proceedings of the SIGCHI conference on human factors in
  computing systems}, pages 21--30. ACM, 2009.

\bibitem{bylinskii2016different}
Z.~Bylinskii, T.~Judd, A.~Oliva, A.~Torralba, and F.~Durand.
\newblock What do different evaluation metrics tell us about saliency models?
\newblock {\em arXiv preprint arXiv:1604.03605}, 2016.

\bibitem{predimportance}
Z.~Bylinskii, N.~W. Kim, P.~O'Donovan, S.~Alsheikh, S.~Madan, H.~Pfister,
  F.~Durand, B.~Russell, and A.~Hertzmann.
\newblock Learning visual importance for graphic designs and data
  visualizations.
\newblock In {\em Proceedings of the 30th Annual ACM Symposium on User
  Interface Software \& Technology}, 2017.

\bibitem{mlnet2016}
M.~Cornia, L.~Baraldi, G.~Serra, and R.~Cucchiara.
\newblock {A Deep Multi-Level Network for Saliency Prediction}.
\newblock In {\em International Conference on Pattern Recognition (ICPR)},
  2016.

\bibitem{cornia2016predicting}
M.~Cornia, L.~Baraldi, G.~Serra, and R.~Cucchiara.
\newblock Predicting human eye fixations via an lstm-based saliency attentive
  model.
\newblock {\em arXiv preprint arXiv:1611.09571}, 2016.

\bibitem{cmo}
S.~Greengard.
\newblock {Mobile Users Say, `It's All About That App, `Bout That App'}.
\newblock \url{http://cmo.cm/2j2tEY2}, 2014.
\newblock [Online; accessed 09-September-2017].

\bibitem{harel2007graph}
J.~Harel, C.~Koch, and P.~Perona.
\newblock Graph-based visual saliency.
\newblock In {\em Advances in neural information processing systems}, pages
  545--552, 2007.

\bibitem{hinton2012improving}
G.~E. Hinton, N.~Srivastava, A.~Krizhevsky, I.~Sutskever, and R.~R.
  Salakhutdinov.
\newblock Improving neural networks by preventing co-adaptation of feature
  detectors.
\newblock {\em arXiv preprint arXiv:1207.0580}, 2012.

\bibitem{hou2012image}
X.~Hou, J.~Harel, and C.~Koch.
\newblock Image signature: Highlighting sparse salient regions.
\newblock {\em IEEE transactions on pattern analysis and machine intelligence},
  34(1):194--201, 2012.

\bibitem{huang2015salicon}
X.~Huang, C.~Shen, X.~Boix, and Q.~Zhao.
\newblock Salicon: Reducing the semantic gap in saliency prediction by adapting
  deep neural networks.
\newblock In {\em Proceedings of the IEEE International Conference on Computer
  Vision}, pages 262--270, 2015.

\bibitem{itti1998model}
L.~Itti, C.~Koch, and E.~Niebur.
\newblock A model of saliency-based visual attention for rapid scene analysis.
\newblock {\em IEEE Transactions on pattern analysis and machine intelligence},
  20(11):1254--1259, 1998.

\bibitem{jacob2003eye}
R.~Jacob and K.~S. Karn.
\newblock Eye tracking in human-computer interaction and usability research:
  Ready to deliver the promises.
\newblock {\em Mind}, 2(3):4, 2003.

\bibitem{judd2009learning}
T.~Judd, K.~Ehinger, F.~Durand, and A.~Torralba.
\newblock Learning to predict where humans look.
\newblock In {\em Computer Vision, 2009 IEEE 12th international conference on},
  pages 2106--2113. IEEE, 2009.

\bibitem{kienzle2007nonparametric}
W.~Kienzle, F.~A. Wichmann, M.~O. Franz, and B.~Sch{\"o}lkopf.
\newblock A nonparametric approach to bottom-up visual saliency.
\newblock In {\em Advances in neural information processing systems}, pages
  689--696, 2007.

\bibitem{dlib09}
D.~E. King.
\newblock Dlib-ml: A machine learning toolkit.
\newblock {\em Journal of Machine Learning Research}, 10:1755--1758, 2009.

\bibitem{kingma2014adam}
D.~P. Kingma and J.~Ba.
\newblock Adam: A method for stochastic optimization.
\newblock In {\em Proceedings of the 3rd International Conference on Learning
  Representations (ICLR)}, 2014.

\bibitem{krafka2016eye}
K.~Krafka, A.~Khosla, P.~Kellnhofer, H.~Kannan, S.~Bhandarkar, W.~Matusik, and
  A.~Torralba.
\newblock Eye tracking for everyone.
\newblock In {\em Proceedings of the IEEE Conference on Computer Vision and
  Pattern Recognition}, pages 2176--2184, 2016.

\bibitem{kruthiventi2017deepfix}
S.~S. Kruthiventi, K.~Ayush, and R.~V. Babu.
\newblock Deepfix: A fully convolutional neural network for predicting human
  eye fixations.
\newblock {\em IEEE Transactions on Image Processing}, 2017.

\bibitem{kummerer2014deep}
M.~K{\"u}mmerer, L.~Theis, and M.~Bethge.
\newblock Deep gaze i: Boosting saliency prediction with feature maps trained
  on imagenet.
\newblock {\em arXiv preprint arXiv:1411.1045}, 2014.

\bibitem{li2016webpage}
J.~Li, L.~Su, B.~Wu, J.~Pang, C.~Wang, Z.~Wu, and Q.~Huang.
\newblock Webpage saliency prediction with multi-features fusion.
\newblock In {\em Image Processing (ICIP), 2016 IEEE International Conference
  on}, pages 674--678. IEEE, 2016.

\bibitem{li2017towards}
Y.~Li, P.~Xu, D.~Lagun, and V.~Navalpakkam.
\newblock Towards measuring and inferring user interest from gaze.
\newblock In {\em Proceedings of the 26th International Conference on World
  Wide Web Companion}, pages 525--533. International World Wide Web Conferences
  Steering Committee, 2017.

\bibitem{liu2016learning}
N.~Liu, J.~Han, T.~Liu, and X.~Li.
\newblock Learning to predict eye fixations via multiresolution convolutional
  neural networks.
\newblock {\em IEEE transactions on neural networks and learning systems},
  2016.

\bibitem{liu2015predicting}
N.~Liu, J.~Han, D.~Zhang, S.~Wen, and T.~Liu.
\newblock Predicting eye fixations using convolutional neural networks.
\newblock In {\em Proceedings of the IEEE Conference on Computer Vision and
  Pattern Recognition}, pages 362--370, 2015.

\bibitem{nair2010rectified}
V.~Nair and G.~E. Hinton.
\newblock Rectified linear units improve restricted boltzmann machines.
\newblock In {\em Proceedings of the 27th international conference on machine
  learning (ICML-10)}, pages 807--814, 2010.

\bibitem{cnnmoney}
J.~O'Toole.
\newblock {Mobile apps overtake PC Internet usage in U.S.}
\newblock \url{http://cnnmon.ie/1hw30fT}, 2014.
\newblock [Online; accessed 09-September-2017].

\bibitem{pan2017salgan}
J.~Pan, C.~Canton, K.~McGuinness, N.~E. O'Connor, J.~Torres, E.~Sayrol, and
  X.~Giro-i Nieto.
\newblock Salgan: Visual saliency prediction with generative adversarial
  networks.
\newblock {\em arXiv preprint arXiv:1701.01081}, 2017.

\bibitem{pan2016shallow}
J.~Pan, E.~Sayrol, X.~Giro-i Nieto, K.~McGuinness, and N.~E. O'Connor.
\newblock Shallow and deep convolutional networks for saliency prediction.
\newblock In {\em Proceedings of the IEEE Conference on Computer Vision and
  Pattern Recognition}, pages 598--606, 2016.

\bibitem{shen2015predicting}
C.~Shen, X.~Huang, and Q.~Zhao.
\newblock Predicting eye fixations on webpage with an ensemble of early
  features and high-level representations from deep network.
\newblock {\em IEEE Transactions on Multimedia}, 17(11):2084--2093, 2015.

\bibitem{shen2014webpage}
C.~Shen and Q.~Zhao.
\newblock Webpage saliency.
\newblock In {\em European Conference on Computer Vision}, pages 33--46.
  Springer, 2014.

\bibitem{still2010saliency}
J.~D. Still and C.~M. Masciocchi.
\newblock A saliency model predicts fixations in web interfaces.
\newblock In {\em 5 th International Workshop on Model Driven Development of
  Advanced User Interfaces (MDDAUI 2010)}, page~25, 2010.

\bibitem{vig2014large}
E.~Vig, M.~Dorr, and D.~Cox.
\newblock Large-scale optimization of hierarchical features for saliency
  prediction in natural images.
\newblock In {\em Proceedings of the IEEE Conference on Computer Vision and
  Pattern Recognition}, pages 2798--2805, 2014.

\bibitem{vincent2010stacked}
P.~Vincent, H.~Larochelle, I.~Lajoie, Y.~Bengio, and P.-A. Manzagol.
\newblock Stacked denoising autoencoders: Learning useful representations in a
  deep network with a local denoising criterion.
\newblock {\em Journal of Machine Learning Research}, 11(Dec):3371--3408, 2010.

\bibitem{xu2014predicting}
J.~Xu, M.~Jiang, S.~Wang, M.~S. Kankanhalli, and Q.~Zhao.
\newblock Predicting human gaze beyond pixels.
\newblock {\em Journal of vision}, 14(1):28--28, 2014.

\bibitem{xu2016spatio}
P.~Xu, Y.~Sugano, and A.~Bulling.
\newblock Spatio-temporal modeling and prediction of visual attention in
  graphical user interfaces.
\newblock In {\em Proceedings of the 2016 CHI Conference on Human Factors in
  Computing Systems}, pages 3299--3310. ACM, 2016.

\bibitem{savvyapps}
K.~Yarmosh.
\newblock {How Much Does an App Cost: A Massive Review of Pricing and other
  Budget Considerations}.
\newblock \url{http://bit.ly/2jLhfHi}, 2017.
\newblock [Online; accessed 09-September-2017].

\bibitem{yu2002color}
H.~Yu, M.~Li, H.-J. Zhang, and J.~Feng.
\newblock Color texture moments for content-based image retrieval.
\newblock In {\em Image Processing. 2002. Proceedings. 2002 International
  Conference on}, volume~3, pages 929--932. IEEE, 2002.

\bibitem{zhang2008sun}
L.~Zhang, M.~H. Tong, T.~K. Marks, H.~Shan, and G.~W. Cottrell.
\newblock Sun: A bayesian framework for saliency using natural statistics.
\newblock {\em Journal of vision}, 8(7):32--32, 2008.

\bibitem{zhao2011learning}
Q.~Zhao and C.~Koch.
\newblock Learning a saliency map using fixated locations in natural scenes.
\newblock {\em Journal of vision}, 11(3):9--9, 2011.

\bibitem{zhao2015saliency}
R.~Zhao, W.~Ouyang, H.~Li, and X.~Wang.
\newblock Saliency detection by multi-context deep learning.
\newblock In {\em Proceedings of the IEEE Conference on Computer Vision and
  Pattern Recognition}, pages 1265--1274, 2015.

\end{thebibliography}

\end{document}